\documentclass{bmvc2k}

\title{Bootstrapping Human Optical Flow and Pose}

\addauthor{Aritro Roy Arko}{aritroroyarko30@gmail.com}{1}
\addauthor{James J. Little}{little@cs.ubc.ca}{1}
\addauthor{Kwang Moo Yi}{kmyi@cs.ubc.ca}{1}

\addinstitution{
 Computer Vision Lab\\
 The University of British Columbia\\
 British Columbia, Canada
}

\runninghead{Arko, Little, Yi}{Bootstrapping Human Optical Flow and Pose}

\def\eg{\emph{e.g}\bmvaOneDot}

\usepackage[inline]{enumitem} %
\usepackage{color}
\usepackage{multirow}
\usepackage[makeroom]{cancel}
\usepackage{placeins} %

\usepackage{wrapfig}

\usepackage{booktabs}
\usepackage[scriptsize]{subfigure}

\definecolor{turquoise}{cmyk}{0.65,0,0.1,0.3}
\definecolor{purple}{rgb}{0.65,0,0.65}
\definecolor{dark_green}{rgb}{0, 0.5, 0}
\definecolor{orange}{rgb}{0.8, 0.6, 0.2}
\definecolor{red}{rgb}{0.8, 0.2, 0.2}
\definecolor{darkred}{rgb}{0.6, 0.1, 0.05}
\definecolor{blueish}{rgb}{0.0, 0.3, .6}
\definecolor{light_gray}{rgb}{0.7, 0.7, .7}
\definecolor{pink}{rgb}{1, 0, 1}
\definecolor{greyblue}{rgb}{0.25, 0.25, 1}

\newcommand{\TODO}[1]{\textbf{\color{red}[TODO: #1]}}

\newcommand{\ky}[1]{{\color{dark_green}#1}}
\newcommand{\KY}[1]{{\color{dark_green}{\bf [KY: #1]}}}
\newcommand{\ar}[1]{{\color{blue}#1}}

\renewcommand{\TODO}[1]{}
\renewcommand{\ky}[1]{#1}
\renewcommand{\KY}[1]{}
\renewcommand{\ar}[1]{}

\newcommand{\Figure}[1]{Figure~\ref{fig:#1}}

\newcommand{\Table}[1]{Table~\ref{tab:#1}}

\newcommand{\Eq}[1]{Eq.~\eqref{eq:#1}}

\newcommand{\Section}[1]{Section~\ref{sec:#1}}

\usepackage{blindtext}

\def \customparskip {.6em}
\renewcommand{\paragraph}[1]{\vspace{\customparskip}\noindent\textbf{#1}}

\usepackage{dsfont}

\newcommand{\loss}[1]{\mathcal{L}_\text{#1}}
\newcommand{\params}[1]{\boldsymbol{\Phi}_\text{#1}}
\newcommand{\flow}{\boldsymbol{\mathcal{F}}}
\newcommand{\IE}{\mathds{E}}
\newcommand{\jointthree}{\mathbf{X}}
\newcommand{\jointtwo}{\mathbf{x}}
\newcommand{\camera}{\mathbf{C}}
\newcommand{\proj}{\mathcal{P}}
\newcommand{\confidence}{w}
\newcommand{\hparam}[1]{\lambda_\text{#1}}
\newcommand{\bx}{\mathbf{x}}

\newcommand{\SupplementaryMaterial}{\texttt{Supplementary Material}\xspace}
\begin{document}
\maketitle
\begin{abstract}
We propose a bootstrapping framework to enhance human optical flow and pose.
We show that, for videos involving humans in scenes, we can improve both the optical flow and the pose estimation quality of humans by considering the two tasks at the same time.
We enhance optical flow estimates by fine-tuning them to fit the human pose estimates and vice versa.
In more detail, we optimize the pose and optical flow networks to, at inference time, agree with each other.
We show that this results in state-of-the-art results on the Human 3.6M and 3D Poses in the Wild datasets, as well as a human-related subset of the Sintel dataset, both in terms of pose estimation accuracy and the optical flow accuracy at human joint locations.
Code available at \url{https://github.com/ubc-vision/bootstrapping-human-optical-flow-and-pose}
\end{abstract}

\section{Introduction}
\label{sec:intro}

Estimating the pose and the motion of humans plays an important role in various tasks in Computer Vision, including human activity recognition~\cite{ullah2018activity}, pedestrian analysis~\cite{gesnouin2020predicting}, and pose-based medical diagnosis~\cite{stenum2021applications}.
Naturally, various methods have been proposed to estimate human pose accurately~\cite{kocabas2020vibe,lin2021end} as well to analyze motion (optical flow) in scenes~\cite{teed2020raft}.
While these methods have been highly successful in the task of human pose estimation and optical flow estimation, respectively, they focus either on solely on the human pose~\cite{kocabas2020vibe,lin2021end}, or focus on generic optical flow~\cite{teed2020raft}.

This leaves room for improvement.
Regarding estimating motions, while unsurprising, indirect evidence of potential for improvement arises from work focusing on human optical flow~\cite{ranjan2018learning,ranjan2020learning}.
Generic optical flow methods, such as Spynet~\cite{ranjan2017optical}, perform better for estimating optical flow of humans when fine-tuned on human-centric scenes.
This demonstrates that when the task revolves around humans, the optical flow method should also focus on humans.
On the other hand, regarding human pose, an important but overlooked assumption in recent works is temporal consistency.
While \cite{hossain2018exploiting} utilized multiple frames to take advantage of temporal consistency, this is, in fact, left for the Neural Network to implicitly figure out and embed into the framework while training. 
However, modern generic optical flow methods perform surprisingly well~\cite{teed2020raft}, so learning this temporal relationship implicitly is a difficulty one does not have to go through---we already have the tools.

\def \figonewidth {0.2}
\begin{figure}[t]
\centering
\subfigure[Pose from \cite{lin2021end}]{\includegraphics[
    width=\figonewidth\linewidth, trim = 300 0 0 0, clip
]{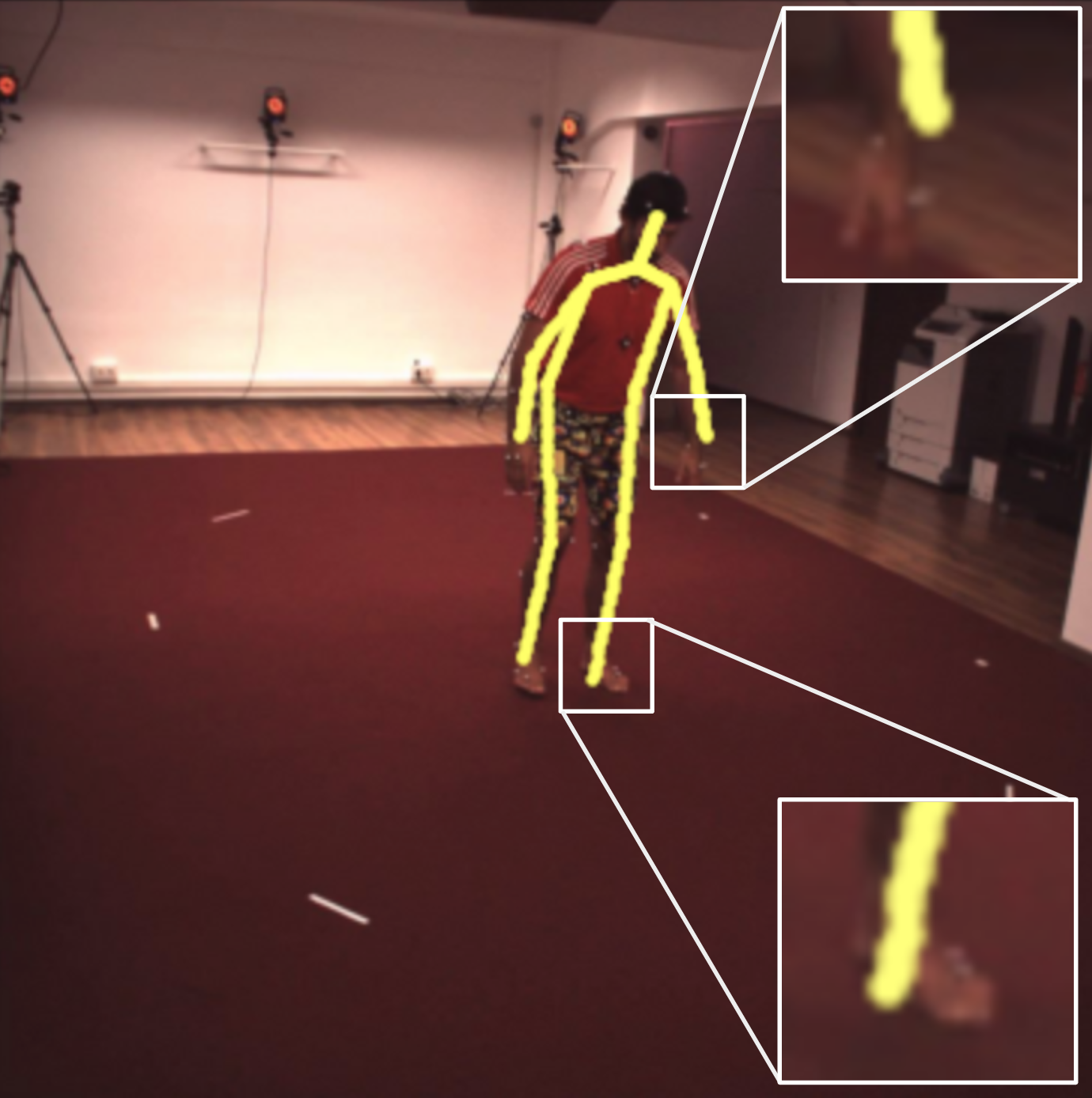}}
\subfigure[Flow from \cite{teed2020raft}]{\includegraphics[
    width=\figonewidth\linewidth, trim = 77 0 0 0, clip
]{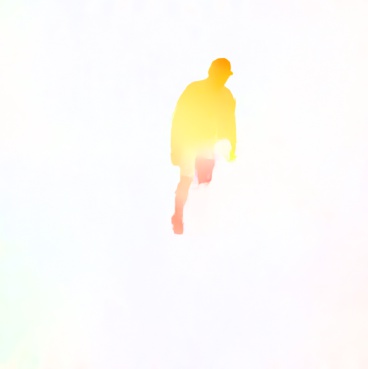}}
\subfigure[Flow enhanced pose]{\includegraphics[
    width=\figonewidth\linewidth, trim = 300 0 0 0, clip
]{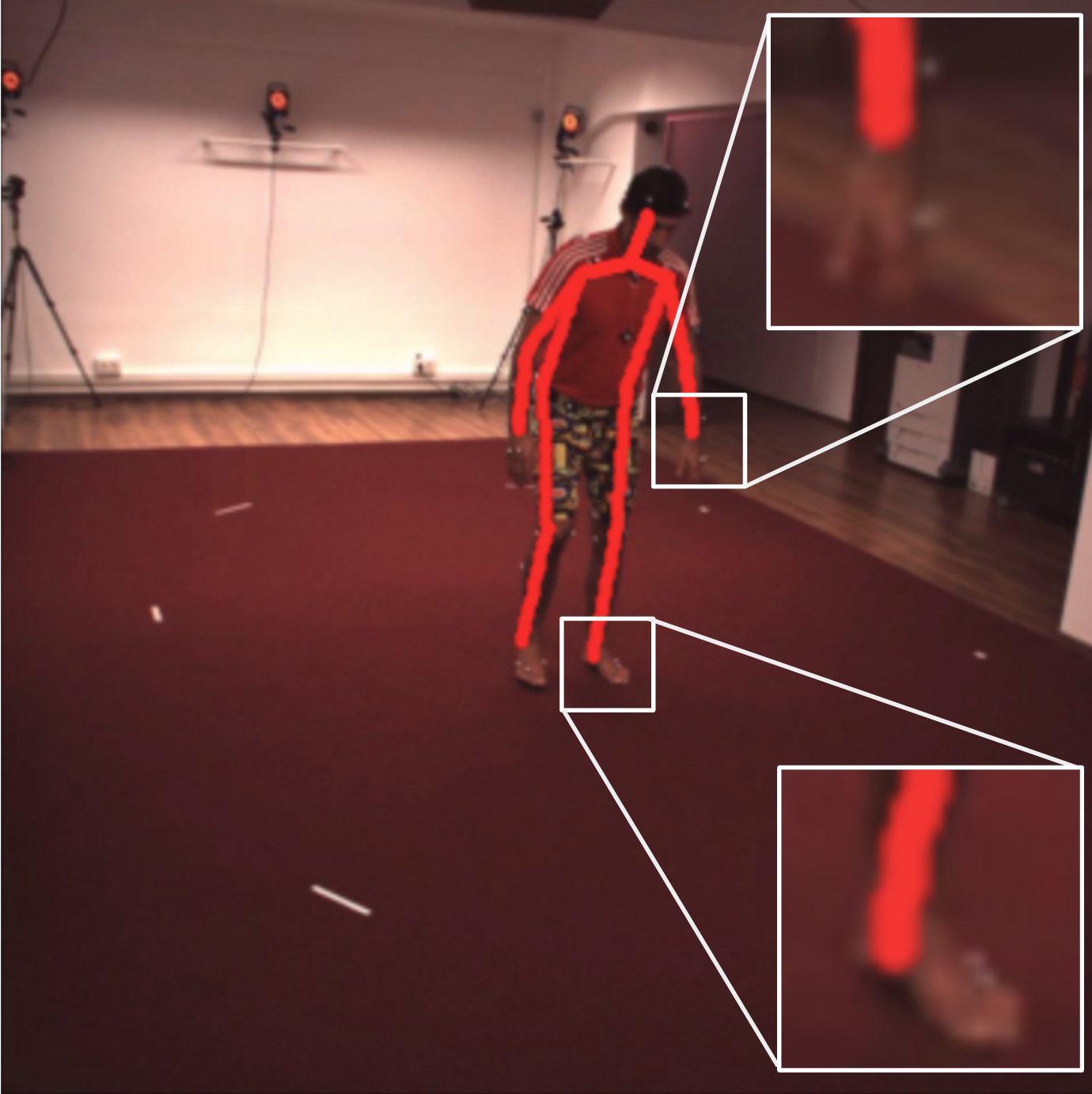}}
\subfigure[Pose enhanced flow]{\includegraphics[
    width=\figonewidth\linewidth, trim = 77 0 0 0, clip
]{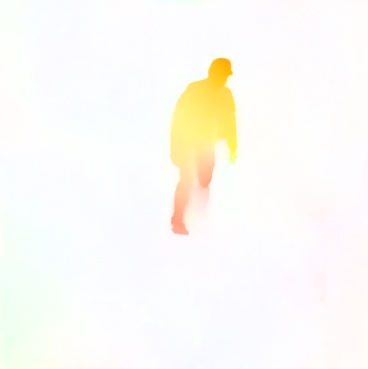}}
\caption{
{\bf Teaser -- }
Given an input image, we use off-the-shelf methods~ \cite{lin2021end}, \cite{teed2020raft} to estimate the pose and the flow of the scene, which we then improve them \emph{without} any retraining.
}
\label{fig:teaser}
\end{figure}
In our work, we propose to make use of the tools that already exist---human pose estimators and optical flow networks---and enhance their performance by marrying the two. 
Our idea originates from the fact that, should human optical flow and pose be estimated properly, they should coincide---the movement of the joints, when projected in 2D, should follow the optical flow estimates at these locations.
We thus create an iterative flow-pose-flow optimization framework for inference, where, \ky{assuming both estimators are not completely failing} we enhance the optical flow to match the pose estimates, which we then optimize the pose to match the flow, and finally optimize the flow once more to match the pose; See~\Figure{teaser} for an example. While this process can be repeated multiple times for further enhancement, with the state-of-the-art methods that are already performing well, we find that this three-round setup is enough for accurate estimates.

In more detail, for the optical flow network we utilize the Recurrent All Pairs Field Transform (RAFT)~\cite{teed2020raft}, and for the human pose estimator the Mesh Transformer (METRO)~\cite{lin2021end}.
We then, given a video sequence of a human, fine-tune the RAFT network so that the flow estimates match the METRO pose estimates at the pixels that correspond to the projected 2D locations of the human skeleton, which would then roughly represent how the human moved between two frames. 
In other words, we create the 2D stick-man representation of the human considering all the bone pixels, and then ensure that the optical flow values at the bone pixels are close to the difference in bone pixel locations in consecutive frames.
With enhanced flow, we obtain enhanced poses by optimizing the 3D poses directly to match the flow, while considering also the temporal smoothness of the estimates, similar to \cite{arnab2019exploiting} and consistent bone length.
Finally, we fine-tune the RAFT network once more with the enhanced poses.
While we initially utilize RAFT and METRO, these can be trivially replaced with any other human pose estimator and optical flow network \ky{as we show in our experiments.}
We note that, during this entire process, no re-training is required and we are strictly fine-tuning to the \textit{test data without any label}.
In other words, we are solving our problem in a setup similar to \textit{transductive} learning.

We validate the efficacy of our method on three datasets: Human 3.6M~\cite{ionescu2013human3} and 3D Poses in the Wild (3DPW)~\cite{von2018recovering} datasets, both of which are human pose datasets, and a subset of the Sintel dataset with scenes containing humans. 
On all three datasets, we show that our method improves METRO and RAFT significantly, achieving state-of-the-art results in terms of both human pose estimation and human optical flow estimation.

To summarize, the main contributions of our work are as follows:
\vspace{-.75em}
\begin{itemize}[leftmargin=*]
\setlength\itemsep{-.5em}
    \item to the best of our knowledge, our work is the first to propose a multi-task-based inference-time optimization framework that enhances both human pose and optical flow estimation;
    \item we achieve the state-of-the-art human pose estimation performance on the Human 3.6M and the 3DPW datasets;
    \item we achieve state-of-the-art human optical flow performance for the Human 3.6M, the 3DPW, and the Sintel (human subset) datasets.
\end{itemize}

\section{Related works}
\label{sec:related}

\paragraph{Monocular 3D human pose estimation.}
Work on monocular 3D human pose estimation can be primarily divided into two categories.
The first class of methods detect 2D keypoints---for example, such as ones provided by Deep Dual Consecutive Network (DCPose)~\cite{liu2021deep}---and then lift them to their 3D counterparts~\cite{martinez2017simple,zhao2017simple,moreno20173d}. While they differ in detail, these lifting processes are typically performed through a Neural Network that is trained from data.

Although lifting-based pose estimators provide highly accurate pose estimates on standard benchmark datasets such as Human3.6M~\cite{ionescu2013human3}, in the case of datasets that are closer to `in-the-wild' setups~\cite{von2018recovering} they do not perform as well.
Moreover, their performance is highly dependent on the quality of the initial detection of 2D keypoints, as if 2D keypoints are wrong, there is often no way of recovery.

Another class of methods utilize a parametric body model, typically the skinned multi-person linear model (SMPL)~\cite{loper2015smpl} and their variants~\cite{SMPL-X:2019,STAR:2020}. Parametric body model-based methods utilize this model in various ways. Bogo et al.~\cite{bogo2016keep} iteratively optimizes the body model parameters to fit the 2D observations. Kanazawa et al.~\cite{kanazawa2018end} regresses the model parameters with a Neural Network, given the input image. VIBE~\cite{kocabas2020vibe} has recently shown that robust video-based pose estimation is possible by incorporating a motion prior learned in an adversarial setup. These methods, however, fail in cases of heavy occlusion, fast motion, and multi-person occlusion, as we will show in our experiments.

Recently for occlusion, Lin et al.~\cite{lin2021end} introduced a transformer-based pose estimation approach (METRO) that performs well on the Human3.6M and 3DPW datasets. 
The main benefit of this method is that, by utilizing transformers, the method learns the relationship among the vertices on the human mesh model, thus the method is able to figure out where the human body parts are, even in the presence of occlusions. Chen et al.~\cite{chen2021anatomy} takes motivation from the anatomy of the human skeleton and break down the task into the bone direction and bone length predictions after which the 3D joint positions are estimated. 
\ky{%
Lastly, ~\cite{pfister2015flowing, cheng2019occlusion, zou2021eventhpe} explore optical flow to improve human pose, but not the other way around.
}%

Tangent to the aforementioned research direction of having a better pose estimator, Arnab et al.~\cite{arnab2019exploiting} focuses on improving what is already available.
They take advantage of the fact that, within neighbouring video frames, multi-view information exists in the presence of camera motions, and enhance pose estimation results via optimizing them according to predefined criteria---matching 2D and 3D pose estimates, enforcing temporal consistency, and utilizing human pose priors.
As our method is also aiming to enhance pose and flow estimates via inference time optimization, these criteria can easily be included, which we do, except for human pose priors since it is often highly data dependent.

\paragraph{Optical flow methods.}
Recent methods use Deep Networks trained on large datasets for estimating optical flow.
Flownet~\cite{dosovitskiy2015flownet}, Flownet~2.0~\cite{ilg2017flownet}, and Spatial Pyramid Network~\cite{ranjan2017optical} use two consecutive frames to estimate optical flow directly in an end-to-end manner, with varying architectural designs to make the algorithm more robust and efficient~\cite{ilg2017flownet} or consider optical flow at various scale levels~\cite{ranjan2017optical}.
On the other hand, Voxel2Voxel~\cite{tran2016deep} utilizes 3D convolutions to take both space and time into account through convolutions, so that more frames than just to two consecutive ones can be taken into account.

In addition to relying on the convolution structure directly, PWC-Net~\cite{Sun2018PWC-Net} utilizes a local cost volume approach, where all pixels in the potentially matching regions are compared with one another, hence explicitly forming relationships that are then utilized to create the optical flow map.
This type of approach, where one allows the Deep Network to explicitly build relationship, has been highly effective in predicting optical flow~\cite{GLUNet_Truong_2020,jiang2021cotr}.
For example, state-of-the-art methods like Recurrent All-Pairs Field Transforms (RAFT)~\cite{teed2020raft} 
and \cite{jiang2021learning}, propose to also utilize cost volumes, together with a recurrent setup that corrects optical flow estimate within the network.
Recently, Correspondence Transformers (COTR)~\cite{jiang2021cotr} further incorporates the cost volume and the recurrent strategy using Transformers~\cite{Vaswani17}.

In case of the performance of these methods on estimating human optical flow, however, is somewhat questionable~\cite{ranjan2018learning,ranjan2020learning}.
Optical flow methods are trained with scenes without particular focus on humans.
These include real-world driving datasets such as KITTI~\cite{geiger2013vision}, and synthetic datasets such as FlyingChairs~\cite{dosovitskiy2015flownet},  FlyingThings~\cite{mayer2016large}, and Sintel~\cite{butler2012naturalistic}.
Because of this, many of them perform poorly when applied to human centred tasks~\cite{ranjan2018learning,ranjan2020learning}.
Thus, there is room for improvement here, which in our work, we bring by utilizing human pose estimators---human optical flow, should human pose estimators be perfect, can be obtained by simply rendering the poses in the two frames from which we wish to extract optical flow.

\section{Method}
\label{sec:method}

\begin{figure}[t]
\begin{center}
\includegraphics[width=0.75\linewidth, trim = 0 0 0 120, clip]{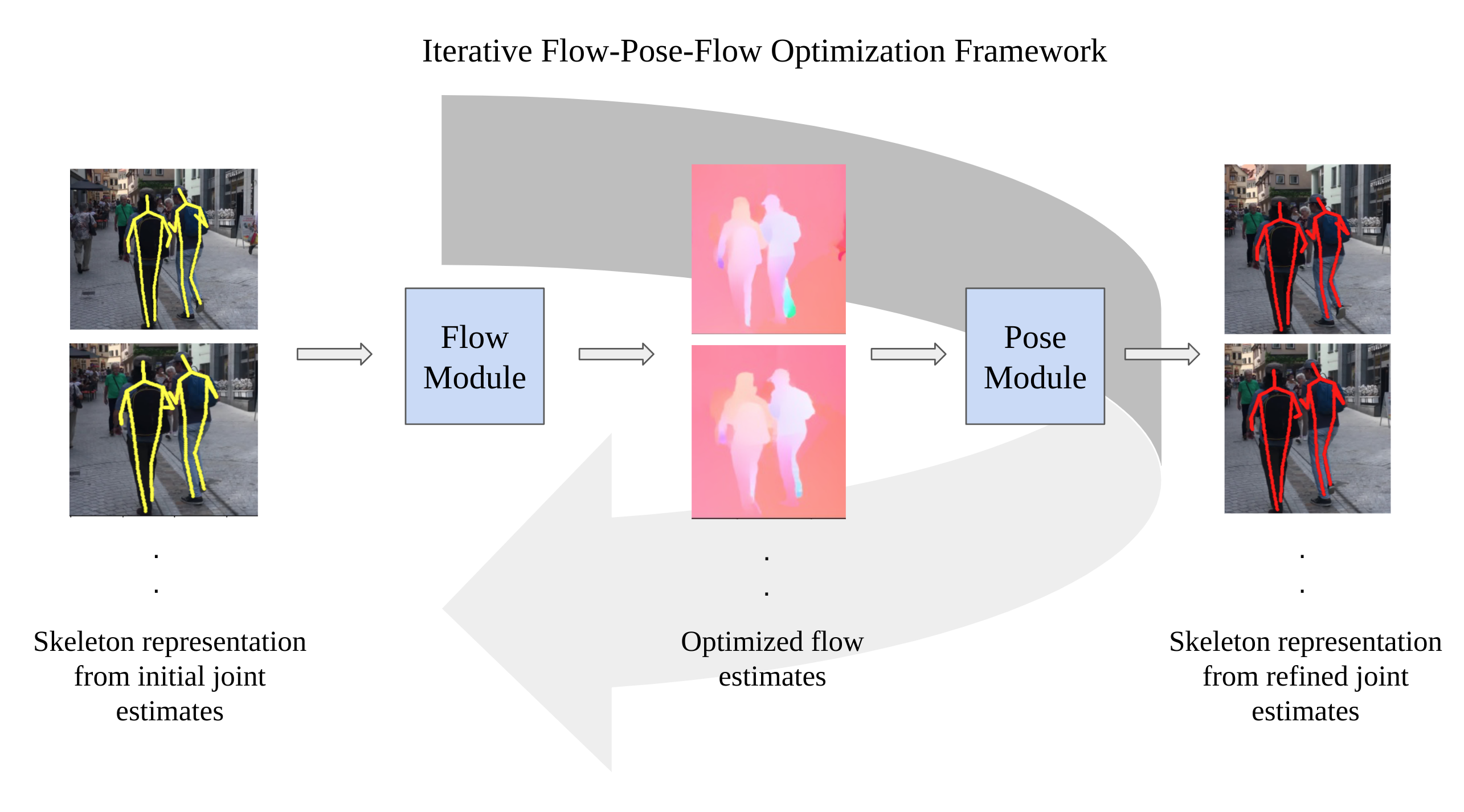}
\vspace{-1em}
\caption{
{\bf Framework -- }
We use off-the-shelf human pose and optical flow estimators and further fine-tune their estimates to coincide with each other for improved performance.
}
\label{fig:framework}
\vspace{-1em}
\end{center}
\end{figure}
The overall framework is shown in \Figure{framework}.
We utilize off-the-shelf pose and flow estimation modules, namely METRO~\cite{lin2021end} for the 3D human pose estimates, DCPose~\cite{liu2021deep} for the location of 2D joints, and RAFT~\cite{teed2020raft} for flow estimation.
With these pre-trained modules, we first obtain the human pose estimates within the scene, with which we create a rough human skeleton-based sketch of the human optical flow.
We then provide this sketch to RAFT, and fine-tune RAFT parameters for each scene so that the estimated flow follows this rough sketch flow. This already results in an improved optical flow estimate compared to simply using RAFT off-the-shelf.
We use this optimized flow estimate and additional human pose related priors to then refine the pose estimates, again by fine-tuning the pose estimates directly on this scene. Finally, this process is repeated with the enhanced estimates. We detail each part of the pipeline in the following subsections.

\subsection{Improving human optical flow with pose}
\label{sec:flow}

\begin{figure}[t]
\centering
\subfigure[Flow from \cite{teed2020raft}]{\includegraphics[width=0.22\linewidth]{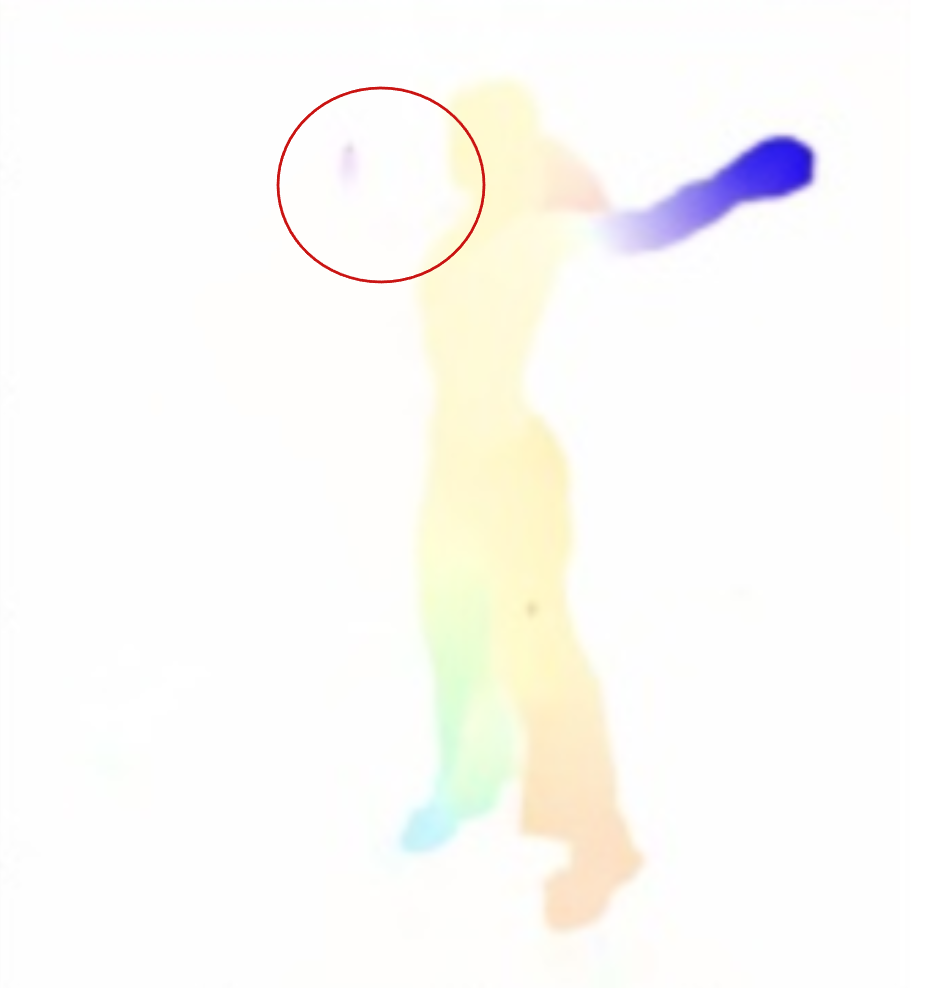}}
\subfigure[Pose-based sketch]{\includegraphics[width=0.22\linewidth]{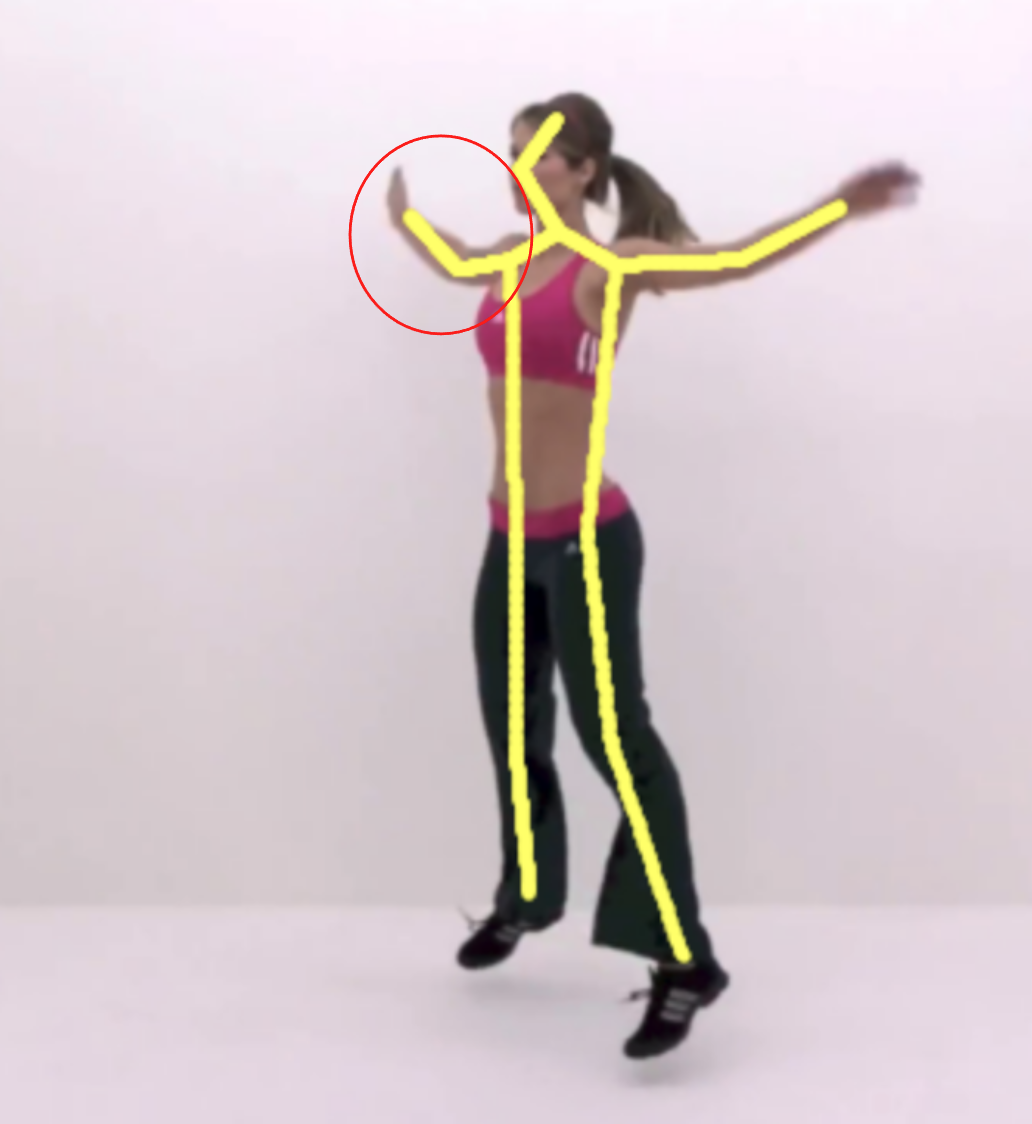}}
\subfigure[Target flow]{\includegraphics[width=0.22\linewidth]{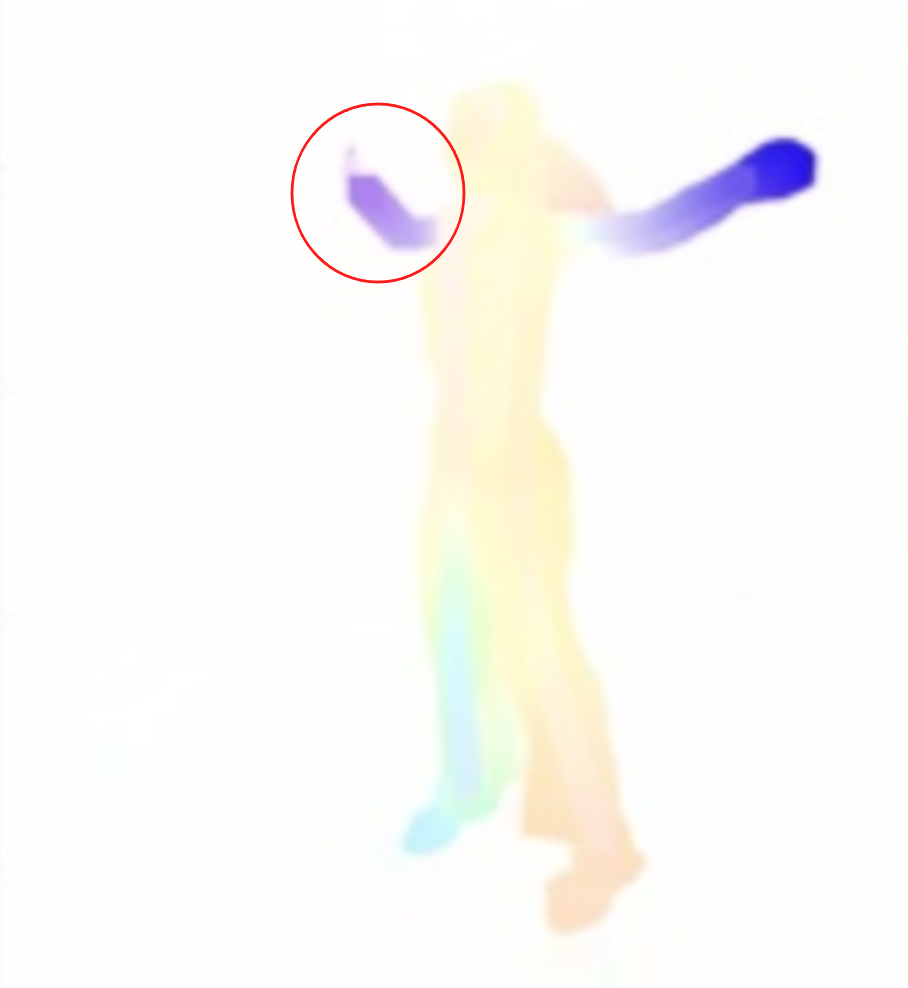}}
\subfigure[Fine-tuned flow]{\includegraphics[width=0.22\linewidth]{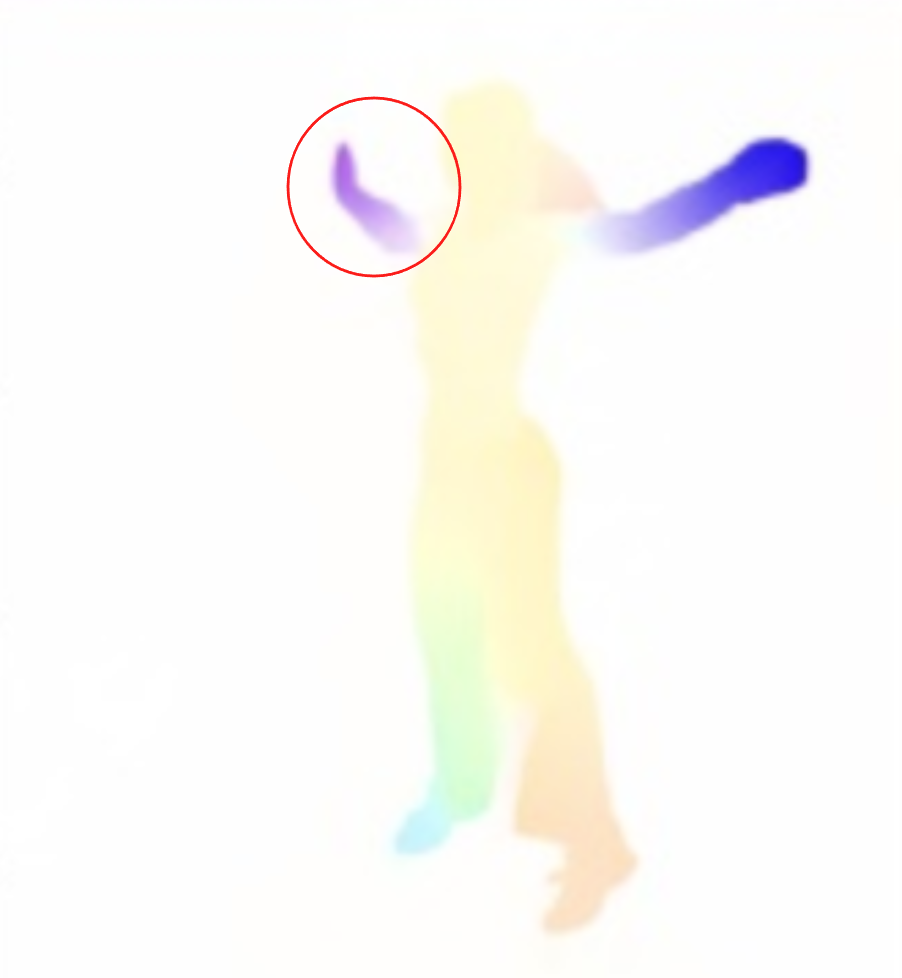}}
\caption{
{\bf Example improving human optical flow -- }
With flow from off-the-shelf RAFT~\cite{teed2020raft} we augment it with the pose estimate-based human optical flow to obtain a target optical flow which we fine-tune RAFT with for improved performance (highlighted circle).
}

\label{fig:flow_sketch}
\end{figure}

To improve human optical flow estimates, we rely on human pose estimation.
As shown in \Figure{flow_sketch}~(a), as RAFT~\cite{teed2020raft} is a generic optical flow estimator, the estimated flow may be inaccurate, especially when a human in the scene is moving abruptly. 
The human pose estimator, for example METRO~\cite{lin2021end}, however, may still provide reasonable (albeit imperfect) pose estimates as shown in the skeleton sketch in \Figure{flow_sketch}~(b).
Hence, allowing the pose estimates to help the optical flow estimation process is a natural choice.

Specifically, for each frame, we apply the 3D human pose estimator, for example, METRO, to get the 3D joint locations.
Next, we project the 3D joints onto the 2D image and create a stick-man figure as shown in \Figure{flow_sketch}~(b), \ky{using the joint pairs as in \cite{liu2021deep} with the exception of the (head, neck) bone being replaced with the (head, nose) and (nose, neck) bones to roughly model also the facial motion.}
\ky{In addition, we make the skeleton thick so that it covers more than just a simple thin line by convolving a cross shape with a fifteen-pixel radius.}
When 3D joint estimates are unreliable, we directly use 2D joint estimates instead, for example from DCPose. We then compute the optical flow on the stick-man pixels trivially by taking the difference between stick-man pixel positions in consecutive frames. This creates a very rough optical flow map of the `bones' of the human body.
Next, the rough optical flow map of the bones is overlaid on top of the estimated flow map (\eg, by RAFT as shown in \Figure{flow_sketch}~(a)). This then leads to a `target' flow map as shown in \Figure{flow_sketch}~(c).

Finally, to leverage the implicit bias of optical flow estimates already stored in RAFT, and at the same time enhance the estimates, we fine-tune the RAFT network to generate this `target' flow map.
As fine-tuning for too long would lead to the RAFT network generating an output identical to the `target' flow map, we fine-tune only for a very few iterations, which leads to an improved optical flow map as shown in \Figure{flow_sketch}~(d).
This somewhat resembles how Deep Image Prior~\cite{ulyanov2018deep} achieves image enhancement via early stopping an overfitting process.
Here, similar to the Deep Image Prior, we are taking advantage of existing networks, and the learned prior about the task within them.
Note that this early stopping also prevents our flow estimates from diverging too much due to potentially faulty estimates.

Mathematically, denoting the flow map prior to the optimization run as $\flow^t$ where $t$ is the frame index, the target flow generated as $\hat{\flow}^t$, smooth $\ell_1$ norm as $\rho$, and the parameters of the RAFT network as $\params{RAFT}$ we minimize
\begin{equation}
\loss{flow}\left(\params{RAFT}\right)=\IE_t\left[\rho\left(\flow^t-\hat{\flow}^t\right)\right]
.
\label{eq:loss_flow}
\end{equation}

\subsection{Improving human pose with optical flow}
\label{sec:pose}

With improved optical flow, we further improve our human pose estimates.
Specifically, we enforce that the movement of the 3D joints, when projected onto the 2D image, follows the optical flow estimates.
In addition to optical flow, we further enforce temporal consistency, as in \cite{arnab2019exploiting}, of the joint and camera estimates, including the human bone length inspired by \cite{yu2021towards}, and also enforce that the results of the 2D joint estimator (DCPose) match that of the 3D method.
Thus, if we denote the 3D joint locations as $\jointthree$ and the corresponding camera estimates as $\camera$, we minimize
\begin{equation}
\begin{split}
\loss{pose}\left(\jointthree, \camera\right) 
= \loss{opt}\left(\jointthree, \camera\right) 
   + \loss{3D}\left(\jointthree\right) 
   + \loss{2D}\left(\jointthree, \camera\right)
   + \loss{temp}\left(\jointthree, \camera\right) 
\label{eq:2}.
\end{split}
\end{equation}

\paragraph{Optical flow consistency---$\loss{opt}\left(\jointthree,\camera\right)$.}
We make sure that the 2D projections of the 3D joints obey the optical flow at the joint locations as the two should be estimating the same phenomenon, just in two different ways.
We thus penalize any difference between the two.
Denoting the projection operation as $\proj$, the optical flow at a pixel location $\bx$ as $\flow_\bx$, and the $j$-th 3D joint location for the $t$-th frame as $\jointthree_j^t$ we write
\begin{equation}
\loss{opt}\left(\jointthree,\camera\right)
=
\hparam{opt}
\IE_{t,j}\left[
\rho
\left(
  \flow^{t-1}_{\proj\left(
    \jointthree_j^{t-1}
  \right)} 
  - 
  \left(
    \proj\left(
      \jointthree_j^t, \camera^t
    \right)
    -
    \proj\left(\jointthree_j^{t-1}, \camera^{t-1}\right)
  \right)
\right)
\right]
.
\end{equation}

\paragraph{3D joint consistency---$\loss{3D}\left(\jointthree\right)$.}
To prevent our optimized poses from deviating too much from the original estimate from the off-the-shelf method, we penalize when the deviation is too large.
Denoting the initial estimate for $\jointthree_j^t$ as $\tilde{\jointthree}_j^t$ we write
\begin{equation}
\loss{3D}\left(\jointthree\right) =
\hparam{3D}
\IE_{t,j}\left[\rho\left(\jointthree_j^t - \tilde{\jointthree}_j^t\right)\right]
.
\end{equation}

\paragraph{2D joint consistency---$\loss{2D}\left(\jointthree,\camera\right)$.}
2D joint estimates from DCPose are typically very accurate, and often more reliable than the 3D pose estimates, which is unsurprising given that estimating 3D pose introduces an additional dimension to the problem.
Hence, we penalize when the 3D estimates deviates from them.
Denoting the detection result from DCPose (or any other 2D human joint detector) as $\jointtwo$, and the projection by $\proj$, we write
\begin{equation}
\loss{2D}\left(\jointthree,\camera\right)
=
\hparam{2D}
\IE_{t,j}\left[
\confidence_j^t
\rho
\left(
\jointtwo_j^t - \proj\left(\jointthree_j^t, \camera^t\right)
\right)
\right]
,
\end{equation}
where $\confidence$ denotes the confidence score for a joint estimate coming from the joint detector.

\paragraph{Temporal consistency---$\loss{temp}\left(\jointthree,\camera\right)$.}
As in \cite{arnab2019exploiting}, we leverage the fact that temporal consistency can be assumed, as change is small between frames. 
Unlike \cite{arnab2019exploiting}, we further enforce the bone length constraint when considering this temporal consistency term.
We thus write
\begin{equation}
\begin{split}
    \loss{temp}\left(\jointthree,\camera\right)
    &=
    \hparam{pos}
    \IE_{t,j}\left[\rho\left(\jointthree_j^t - \jointthree_j^{t-1}\right)\right]
    +
    \hparam{cam}
    \IE_{t,j}\left[\rho\left(\camera^t - \camera^{t-1}\right)\right]
    \\
    &\qquad+
    \hparam{bone}
    \IE_{t,j,k}\left[
    \rho\left(
    \left\|\jointthree_j^t - \jointthree_k^t\right\|_2
    -
    \left\|\jointthree_j^{t-1} - \jointthree_k^{t-1}\right\|_2
    \right)\right]
\end{split}
,
\end{equation}
where $\|\cdot\|_2$ denotes the $\ell_2$ norm.

\subsection{Implementation details}
\label{sec:implem_details}
\paragraph{Optimization settings.}
To optimize the flow network, we use the Adam optimizer~\cite{kingma2014adam} with a learning rate of $10^{-5}$ and default parameters. 
For the flow network (RAFT), we optimize its parameters for the entire video, per \Eq{loss_flow}.
This effectively results in an improved RAFT model for each scene.
In more detail, we iterate over each frame one-by-one (equivalent to batch size of one) and optimize the per-scene RAFT network for eight epochs, a value that we empirically found that works well in general.

For optimizing the 3D joint estimates, we, again, use the Adam optimizer~\cite{kingma2014adam} with a learning rate of 0.001.
We empirically set the number of optimization iterations to 1,500 epochs.
We perform this flow and pose optimization cycle once for pose and twice for flow.
See \SupplementaryMaterial for experiments regarding this choice.

Finally, for our experiments with METRO, we set
$\hparam{opt}=0.01$,
$\hparam{3D}=400$,
$\hparam{2D}=0.01$,
$\hparam{pos}=300$,
$\hparam{cam}=0.1$, and
$\hparam{bone}=10^4$,
which we found empirically by testing a few videos that this setup works well in general. 

\paragraph{When 3D joint estimates are unreliable.}
We found that METRO, our choice of the 3D joint estimator, does not perform as well when applied to datasets that are not focusing on humans. 
This leads into $\jointthree$ being erroneous.
In this case, we simply resort to the 2D joint detector DCPose, which delivers more robust performance in more complicated scenarios.
We thus modify our losses, specifically directly replace the projected points $\proj\left(\jointthree_j^{t}, \camera^{t}\right)$ with $\jointtwo^t$ for all losses.
We also ignore loss components that directly use $\jointthree$ and not $\proj\left(\jointthree_j^{t}, \camera^{t}\right)$ and replace the 3D joint consistency with a 2D version using initial 2D estimates.
Specifically, this setup is for the Sintel Human subset dataset, which we will discuss more in \Section{sintel}.
With this setup, we optimize for 50 epochs per cycle  for the flow instead of the eight previously.
The setup for the joints remain the same, although we are now in 2D.

\section{Results}
\label{sec:results}
For evaluation, we use the Human3.6M~\cite{ionescu2013human3}, 3DPW~\cite{von2018recovering} and a subset of the Sintel (final)~\cite{butler2012naturalistic} datasets.
In this section, we show that our method outperforms the state of the art human pose and optical flow results on all the three datasets, without introducing any retraining or additional datasets.

\subsection{Datasets and evaluation setup}
\label{sec:data}

\vspace{-\customparskip}
\paragraph{Datasets.} 
We evaluate both the performance of human pose estimation and human optical flow estimation on three datasets---two aimed at human pose estimation and another at optical flow.
\vspace{-.75em}
\begin{itemize}[leftmargin=*]
\setlength\itemsep{-.5em}
  \item \textbf{Human3.6M~\cite{ionescu2013human3}:}
  This dataset is a large scale human pose dataset with 2D and 3D annotations captured in an indoor setting.
  There are a total of 2 subjects, S9 and S11, performing different actions such as walking and sitting. 
  We follow the exact same evaluation setup as in METRO~\cite{lin2021end}, based on their public implementation.
  As in \cite{lin2021end} and \cite{pavlakos2017coarse}, we down-sample the videos from 50 fps to 10 fps.
  \item \textbf{3DPW~\cite{von2018recovering}:} 
  This dataset is another large-scale dataset with 3D annotations of people \emph{in the wild}.
  We use the standard test set of the 3DPW dataset---a total of 35K frames.
  \item \textbf{Sintel human subset:} 
  \label{sec:sintel}
  We create a subset of the Sintel (final) dataset~\cite{butler2012naturalistic}---a commonly used optical flow dataset of synthetic scenes with atmospheric effects---by selecting scenes with humans in  them: \ky{`alley 2', `bamboo 2', `cave 2', and `cave 4'}.
This results in a small dataset consisting of 196 frames with which we evaluate the human optical flow.
\end{itemize}

\paragraph{Metrics.} 
We evaluate both the joint positions and the human optical flow errors with standard metrics.
We use mean per joint position error (MPJPE) and the end point error (EPE).
For the Human3.6M dataset and the 3DPW dataset we report the EPE values only for the joint locations where ground-truth correspondences are available---we do not know the ground-truth flow for all other points. For the Sintel dataset we use all points. In addition, we follow the standard method of computing MPJPE for the fourteen common joints.

\begin{figure}
\centering
\subfigure[METRO]{\includegraphics[width=0.115\linewidth]{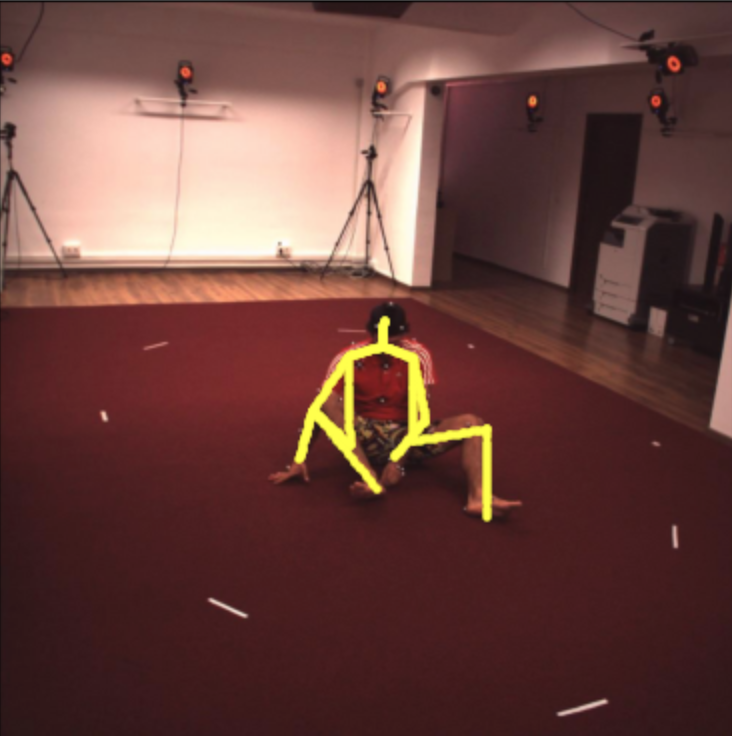}}
\subfigure[Ours]{\includegraphics[width=0.115\linewidth]{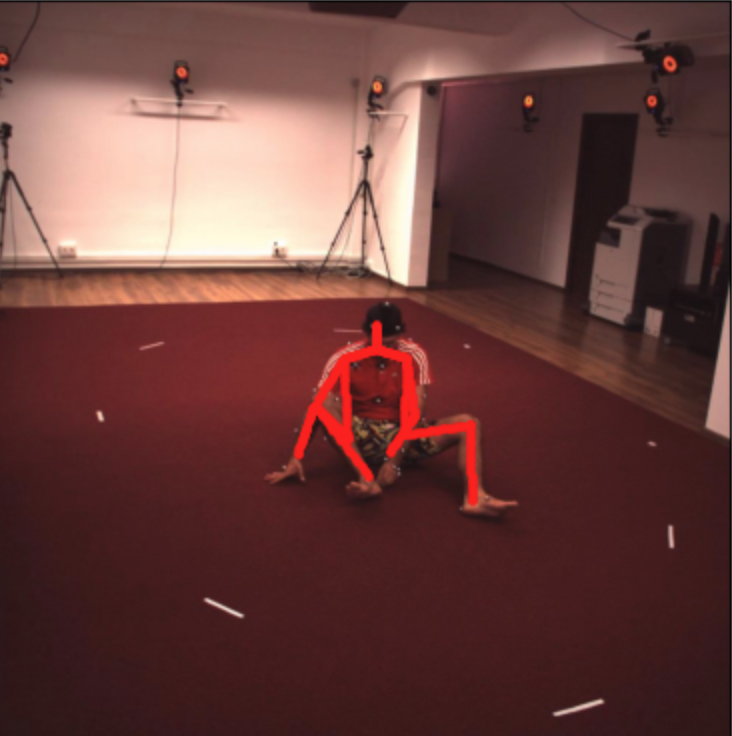}}
\subfigure[METRO]{\includegraphics[width=0.115\linewidth]{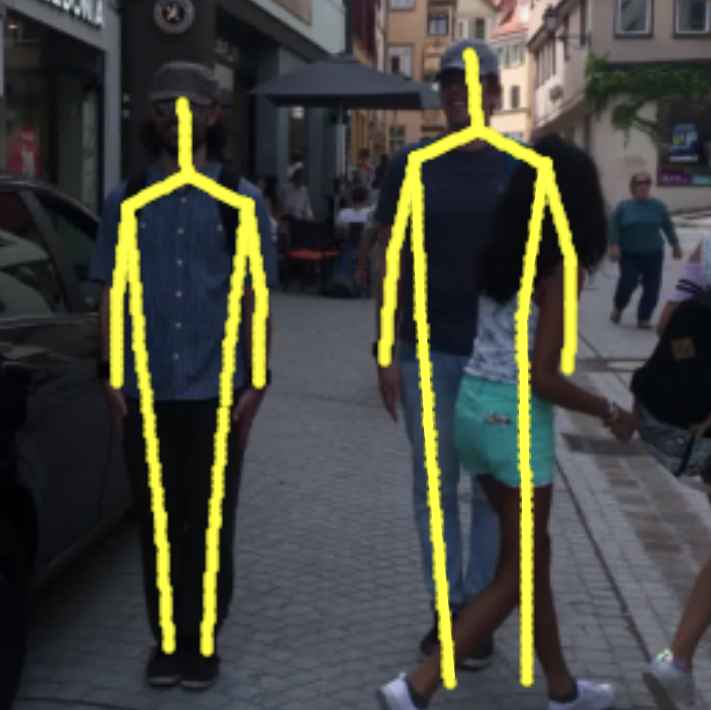}}
\subfigure[Ours]{\includegraphics[width=0.115\linewidth]{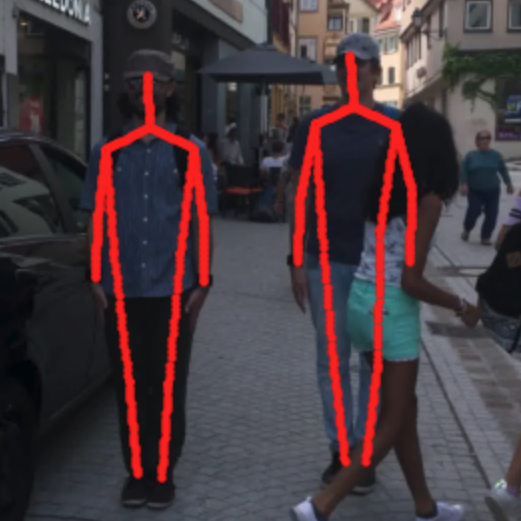}}
\subfigure[DCPose]{\includegraphics[width=0.115\linewidth]{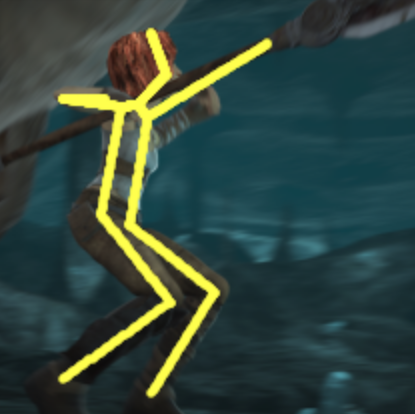}}
\subfigure[Ours]{\includegraphics[width=0.115\linewidth]{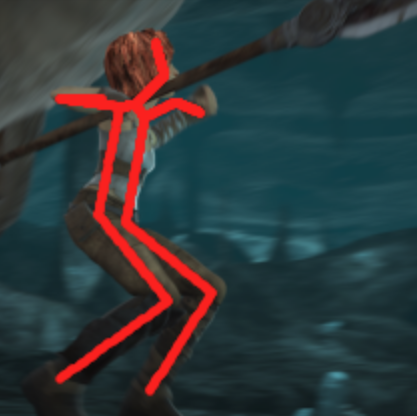}}
\subfigure[DCPose]{\includegraphics[width=0.115\linewidth]{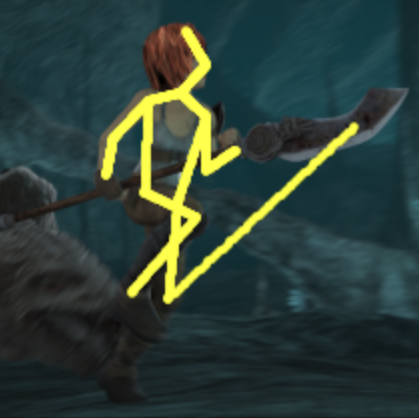}}
\subfigure[Ours]{\includegraphics[width=0.115\linewidth]{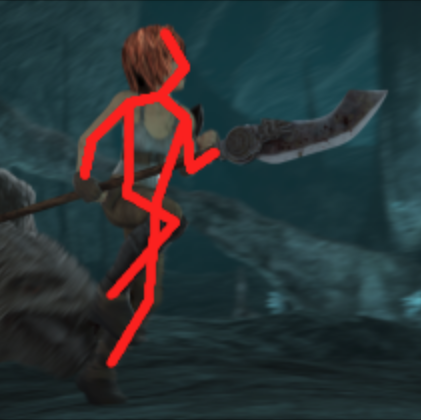}}

\caption{
{\bf Pose estimation examples -- }
Example comparison of pose estimation results on the (a--b) Human3.6m, (c--d) on the 3DPW, and (e--h) on the Sintel human subset dataset.
Note how our method improves pose estimates in case of occlusions (c--d) or when the person is interacting with the object (e--h).
}%
\label{fig:qualitative_pose}
\end{figure}

\begin{figure}
\centering

\subfigure[Input]{\includegraphics[
    width=0.115\linewidth, trim=200 0 200 0, clip
]{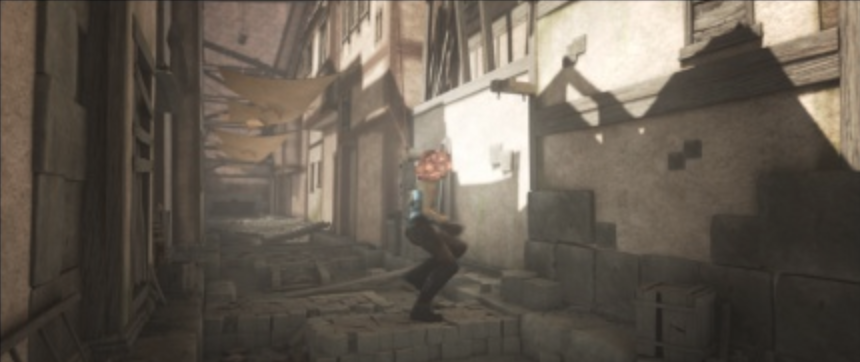}}
\subfigure[GT]{\includegraphics[
    width=0.115\linewidth, trim=200 0 200 0, clip
]{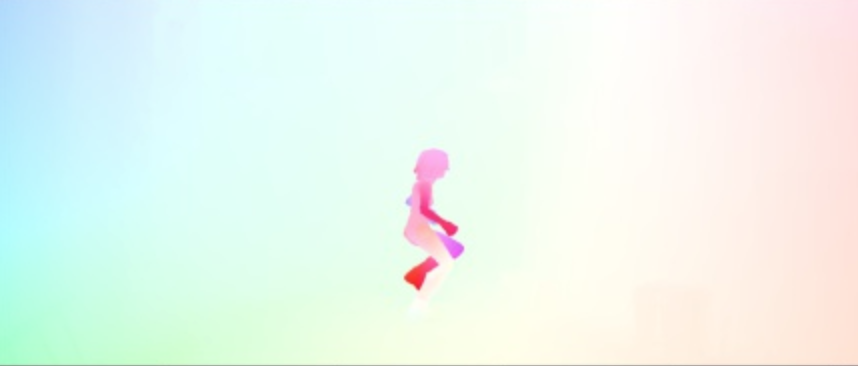}}
\subfigure[RAFT \cite{teed2020raft}]{\includegraphics[
    width=0.115\linewidth, trim=200 0 200 0, clip
]{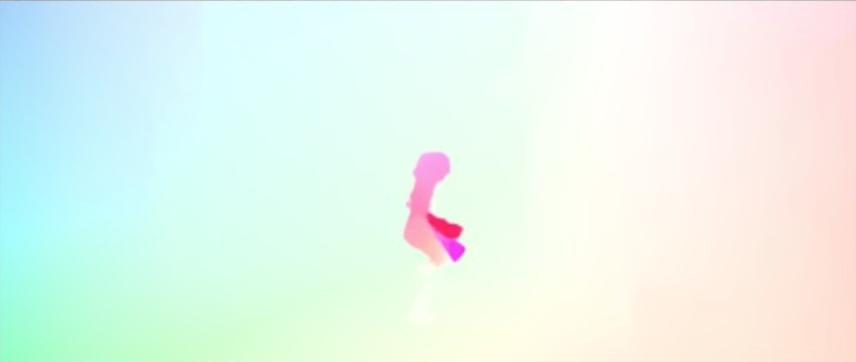}}
\subfigure[Ours]{\includegraphics[
    width=0.115\linewidth, trim=200 0 200 0, clip
]{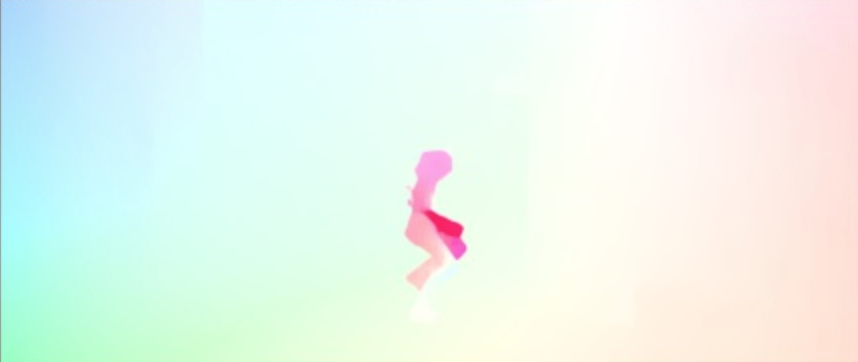}}
\subfigure[Input]{\includegraphics[
    width=0.115\linewidth, trim=200 0 200 0, clip
]{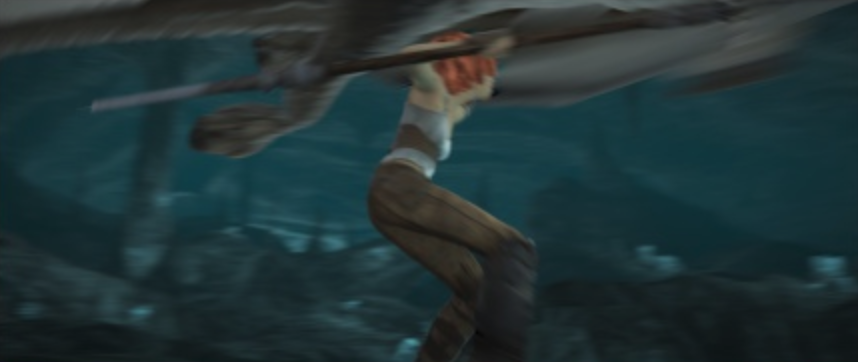}}
\subfigure[GT]{\includegraphics[
    width=0.115\linewidth, trim=200 0 200 0, clip
]{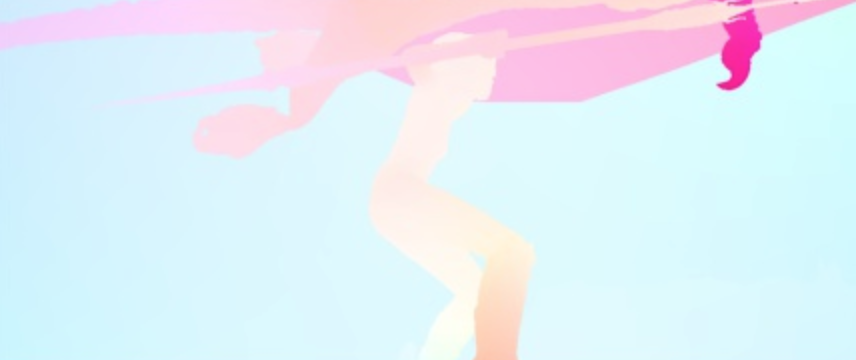}}
\subfigure[RAFT \cite{teed2020raft}]{\includegraphics[
    width=0.115\linewidth, trim=200 0 200 0, clip
]{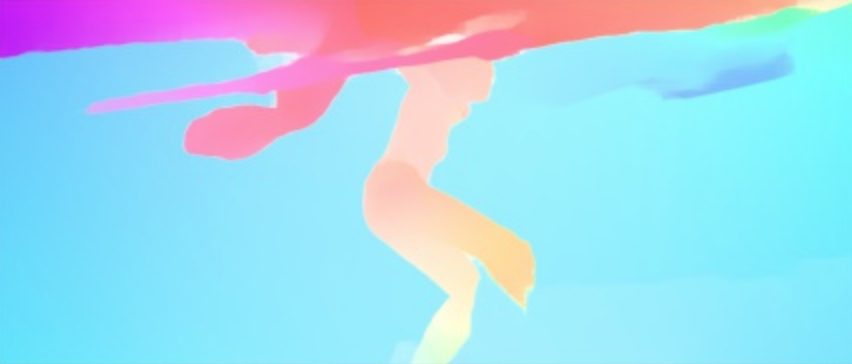}}
\subfigure[Ours]{\includegraphics[
    width=0.115\linewidth, trim=200 0 200 0, clip
]{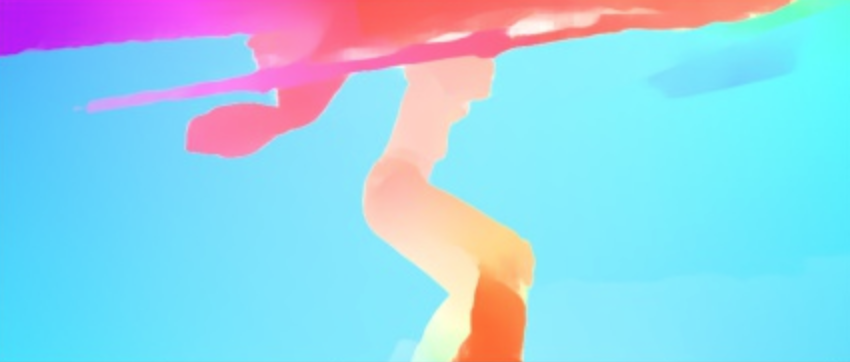}}
\caption{
{\bf Flow estimation examples -- }
Example comparison of optical flow estimation results on the Sintel human subset dataset. 
Ours successfully recovers the left leg in (d) and the right leg in (h), which was wrongly estimated in the case of the original RAFT method.
}%
\label{fig:qualitative_flow}
\end{figure}

\subsection{Experimental results}

\paragraph{Qualitative results.}
In \Figure{qualitative_pose}, we show qualitative results for how human pose estimation improves with our method.
Note how the pose estimation result of existing methods can lead to partial inaccuracies, for example, due to  (a--b) complex poses, (c--d) occlusions with other entities in the scene and (e--h) object interactions.
Our method provides accurate results even in these cases.
In \Figure{qualitative_flow}, we show results for improving flow estimates.
As shown, optical flow values on the human that were originally missing and wrongly estimated are corrected with our method.

\begin{table}
    \begin{minipage}[t]{0.56\linewidth}
    \begin{center}
    \resizebox{\linewidth}{!}{
        \setlength{\tabcolsep}{8pt}
        \begin{tabular}{@{}l cc cc cc@{}} 
         \toprule
          & \multicolumn{2}{c}{Human3.6M} & \multicolumn{2}{c}{3DPW} & \multicolumn{2}{c}{Sintel human}\\ 
          \cmidrule(lr){2-3} \cmidrule(lr){4-5} \cmidrule(lr){6-7}
         Model & MPJPE$\downarrow$ & EPE$\downarrow$ & MPJPE $\downarrow$& EPE $\downarrow$& MPJPE $\downarrow$& EPE $\downarrow$\\ 
         \midrule
         HMR \cite{kanazawa2018end} & 88.00 & - & - & - & - & -\\
         Pose2Mesh \cite{choi2020pose2mesh} & 64.90 & - & 89.20 & - & - & -\\
         I2LMeshNet \cite{moon2020i2l} & 64.90 & - & 93.20 & - & - & -\\
         VIBE \cite{kocabas2020vibe} &  65.60 & - & 82.00 & - & - & -\\
         PARE \cite{kocabas2021pare} &  - & - & 74.50 & - & - & -\\
         METRO \cite{lin2021end} & 54.04 & - & 77.10 & - & - & -\\
         METRO$^\ast$ & 54.07 & - & 75.87 & - & - & -\\
         
         \midrule
         RAFT \cite{teed2020raft} & - & 2.729 & - & 1.751 & - & 2.513\\
         
         \midrule
         Our method & \textbf{53.15} & \textbf{2.402} & \textbf {74.45} & \textbf {1.661} & - & \textbf {2.383}\\
         
         \bottomrule
        \end{tabular}
    }
    \end{center}
    \caption{
    {\bf Quantitative results -- }
    Comparison with other 3D pose and optical flow methods multiple datasets. 
    METRO$^\ast$ represents our adaptation of \cite{lin2021end} based on the official public implementation.
    Our method allows improving upon the state of the art, both in terms of human optical flow and pose.
    }
    \label{tab:quantitative}
    \end{minipage}
    \hfill
    \begin{minipage}[t]{0.415\linewidth}
    \vspace{-4.8em} %
    \begin{center}
    \resizebox{\linewidth}{!}{
    \setlength{\tabcolsep}{2pt}
    \begin{tabular}{@{}l cc @{}} 
     \toprule
      & \multicolumn{2}{c}{Human3.6M} \\ 
      \cmidrule(lr){2-3} 
     Model & MPJPE$\downarrow$ & EPE$\downarrow$ \\ 
     \midrule
     METRO & 54.04 & - \\
     METRO$^\ast$ & 54.07 & - \\
     Anatomy3D & 47.90 & -\\
     Avg(Anatomy3D, METRO)  & 43.34 & - \\
     \midrule
     RAFT & - & 2.729 \\
     GMA  & - & 2.740 \\
     Avg(GMA, RAFT)  & - & 2.688 \\
     \midrule
     Our method -- METRO / RAFT & 53.15 &  2.402 \\
     Our method -- METRO / GMA & 53.14 &  2.316 \\
     Our method -- Anatomy3D / RAFT & 46.93 & 2.324 \\
     Our method -- Anatomy3D / GMA & 46.93 & 2.250 \\
     Our method -- Avg(Anatomy3D, METRO)/Avg(GMA, RAFT)  & \textbf{42.56} & \textbf{2.230}  \\
     \bottomrule
    \end{tabular}
    }
    \end{center}
    \caption{
    \textbf{Refinement on top of different 3D pose and optical flow methods} – A performance comparison on the Human3.6M dataset showing that our method is applicable to various methods.
    }%
    \label{tab:quantitative_extra}
    \end{minipage}
\end{table}

\paragraph{Quantitative results.}
In \Table{quantitative}, we report the MPJPE and the EPE metrics for our method as well as other baselines.
Building upon METRO and RAFT, our method outperforms both methods in their respective tasks.
It is worth noting that, as shown earlier in \Figure{qualitative_pose}, the gains are most prominent when extreme poses occur and occlusions exist.
Still, they are frequent and large enough to be seen in the average metrics that we report in this table.

To further show that our optimization framework is not limited to METRO and RAFT, we conduct additional experiments with other optical flow and human pose estimators: GMA~\cite{jiang2021learning}, Anatomy3D~\cite{chen2021anatomy}.
We also use them together with RAFT and METRO by averaging their estimates, which we found to work well.

We note that Anatomy3D does not provide camera estimates, and is designed to utilize the ground-truth camera parameters. 
As such, when Anatomy3D is used, we rely on ground-truth cameras.
In addition, Anatomy3D estimates are initially more accurate than those from METRO, rendering the temporal consistency term useless.
We thus do not use this term when Anatomy3D estimates are used.

We report these results in the \Table{quantitative_extra}---our method enhances \emph{all} methods, demonstrating its efficacy.

\paragraph{Ablation study.}
We provide an ablation study on the losses and the number of optimization cycles in the \SupplementaryMaterial.

\section{Conclusions}

We proposed an iterative framework for enhancing human pose and optical flow estimation accuracy of existing methods \emph{without} any training.
Our method takes its roots from the fact that, when performed properly, the two tasks should coincide.
Hence, we optimize optical flow to match the movement of the human joints and vice versa.
This leads to a performance boost, enough to push the boundaries of the state of the art further.
The gain was especially visible in cases where extreme poses, occlusions, and object-human interactions exist. We have validated our method on two human pose datasets, Human3.6M and 3DPW, and a subset of the Sintel optical flow dataset, achieving state of the art in all three datasets, both in terms of human pose and optical flow estimation accuracy.

\paragraph{Limitations and future work.}
The performance of our framework, while it improves upon the state of the art, is also somewhat bound by the quality of the initial estimates.
Hence, starting completely from a wrong pose would not lead to accurate pose estimates, even with good optical flow.
The opposite is also true.
Thus, a promising research direction would be to include multiple methods into the framework, thus reducing the risk of total failure.
A naive version of this was shown in \Table{quantitative_extra} via averaging, but a more sophisticated way of combining methods, for example via a multiple-instance learning setup could be interesting.

\ky{%
Additionally, the inference time of our pipeline is a limitation as we are, in fact, optimizing a network during inference time---it takes 20.50 seconds per frame on a GeForce RTX 3090 GPU using METRO and RAFT. 
This is roughly 14.5 times slower than running METRO and RAFT separately. 
Speeding up this inference time could be an interesting future research direction.
}%
 
\bibliography{main}

\begin{thebibliography}{44}
\providecommand{\natexlab}[1]{#1}
\providecommand{\url}[1]{\texttt{#1}}
\expandafter\ifx\csname urlstyle\endcsname\relax
  \providecommand{\doi}[1]{doi: #1}\else
  \providecommand{\doi}{doi: \begingroup \urlstyle{rm}\Url}\fi

\bibitem[Arnab et~al.(2019)Arnab, Doersch, and Zisserman]{arnab2019exploiting}
Anurag Arnab, Carl Doersch, and Andrew Zisserman.
\newblock Exploiting temporal context for 3d human pose estimation in the wild.
\newblock In \emph{Proceedings of the IEEE/CVF Conference on Computer Vision
  and Pattern Recognition}, pages 3395--3404, 2019.

\bibitem[Bogo et~al.(2016)Bogo, Kanazawa, Lassner, Gehler, Romero, and
  Black]{bogo2016keep}
Federica Bogo, Angjoo Kanazawa, Christoph Lassner, Peter Gehler, Javier Romero,
  and Michael~J Black.
\newblock Keep it smpl: Automatic estimation of 3d human pose and shape from a
  single image.
\newblock In \emph{European conference on computer vision}, pages 561--578.
  Springer, 2016.

\bibitem[Butler et~al.(2012)Butler, Wulff, Stanley, and
  Black]{butler2012naturalistic}
Daniel~J Butler, Jonas Wulff, Garrett~B Stanley, and Michael~J Black.
\newblock A naturalistic open source movie for optical flow evaluation.
\newblock In \emph{European conference on computer vision}, pages 611--625.
  Springer, 2012.

\bibitem[Chen et~al.(2021)Chen, Fang, Shen, Zhu, Chen, and
  Luo]{chen2021anatomy}
Tianlang Chen, Chen Fang, Xiaohui Shen, Yiheng Zhu, Zhili Chen, and Jiebo Luo.
\newblock {Anatomy-Aware 3D Human Pose Estimation with Bone-Based Pose
  Decomposition}.
\newblock \emph{IEEE Trans. Circuits Syst. Video Technol.}, 32\penalty0
  (1):\penalty0 198--209, 2021.

\bibitem[Cheng et~al.(2019)Cheng, Yang, Wang, Yan, and Tan]{cheng2019occlusion}
Yu~Cheng, Bo~Yang, Bo~Wang, Wending Yan, and Robby~T Tan.
\newblock Occlusion-aware networks for 3d human pose estimation in video.
\newblock In \emph{Proceedings of the IEEE/CVF international conference on
  computer vision}, pages 723--732, 2019.

\bibitem[Choi et~al.(2020)Choi, Moon, and Lee]{choi2020pose2mesh}
Hongsuk Choi, Gyeongsik Moon, and Kyoung~Mu Lee.
\newblock Pose2mesh: Graph convolutional network for 3d human pose and mesh
  recovery from a 2d human pose.
\newblock In \emph{European Conference on Computer Vision}, pages 769--787.
  Springer, 2020.

\bibitem[Dosovitskiy et~al.(2015)Dosovitskiy, Fischer, Ilg, Hausser, Hazirbas,
  Golkov, Van Der~Smagt, Cremers, and Brox]{dosovitskiy2015flownet}
Alexey Dosovitskiy, Philipp Fischer, Eddy Ilg, Philip Hausser, Caner Hazirbas,
  Vladimir Golkov, Patrick Van Der~Smagt, Daniel Cremers, and Thomas Brox.
\newblock Flownet: Learning optical flow with convolutional networks.
\newblock In \emph{ICCV}, pages 2758--2766, 2015.

\bibitem[Geiger et~al.(2013)Geiger, Lenz, Stiller, and
  Urtasun]{geiger2013vision}
Andreas Geiger, Philip Lenz, Christoph Stiller, and Raquel Urtasun.
\newblock Vision meets robotics: The kitti dataset.
\newblock \emph{The International Journal of Robotics Research}, 32\penalty0
  (11):\penalty0 1231--1237, 2013.

\bibitem[Gesnouin et~al.(2020)Gesnouin, Pechberti, Bresson, Stanciulescu, and
  Moutarde]{gesnouin2020predicting}
Joseph Gesnouin, Steve Pechberti, Guillaume Bresson, Bogdan Stanciulescu, and
  Fabien Moutarde.
\newblock Predicting intentions of pedestrians from 2d skeletal pose sequences
  with a representation-focused multi-branch deep learning network.
\newblock \emph{Algorithms}, 13\penalty0 (12):\penalty0 331, 2020.

\bibitem[Hossain and Little(2018)]{hossain2018exploiting}
Mir Rayat~Imtiaz Hossain and James~J Little.
\newblock Exploiting temporal information for 3d human pose estimation.
\newblock In \emph{Proceedings of the European Conference on Computer Vision
  (ECCV)}, pages 68--84, 2018.

\bibitem[Ilg et~al.(2017)Ilg, Mayer, Saikia, Keuper, Dosovitskiy, and
  Brox]{ilg2017flownet}
Eddy Ilg, Nikolaus Mayer, Tonmoy Saikia, Margret Keuper, Alexey Dosovitskiy,
  and Thomas Brox.
\newblock Flownet 2.0: Evolution of optical flow estimation with deep networks.
\newblock In \emph{Proceedings of the IEEE conference on computer vision and
  pattern recognition}, pages 2462--2470, 2017.

\bibitem[Ionescu et~al.(2013)Ionescu, Papava, Olaru, and
  Sminchisescu]{ionescu2013human3}
Catalin Ionescu, Dragos Papava, Vlad Olaru, and Cristian Sminchisescu.
\newblock Human3. 6m: Large scale datasets and predictive methods for 3d human
  sensing in natural environments.
\newblock \emph{IEEE transactions on pattern analysis and machine
  intelligence}, 36\penalty0 (7):\penalty0 1325--1339, 2013.

\bibitem[Jiang et~al.(2021{\natexlab{a}})Jiang, Campbell, Lu, Li, and
  Hartley]{jiang2021learning}
Shihao Jiang, Dylan Campbell, Yao Lu, Hongdong Li, and Richard Hartley.
\newblock {Learning to Estimate Hidden Motions with Global Motion Aggregation}.
\newblock In \emph{CVPR}, 2021{\natexlab{a}}.

\bibitem[Jiang et~al.(2021{\natexlab{b}})Jiang, Trulls, Hosang, Tagliasacchi,
  and Yi]{jiang2021cotr}
Wei Jiang, Eduard Trulls, Jan Hosang, Andrea Tagliasacchi, and Kwang~Moo Yi.
\newblock {COTR: Correspondence Transformer for Matching Across Images}.
\newblock In \emph{ICCV}, 2021{\natexlab{b}}.

\bibitem[Kanazawa et~al.(2018)Kanazawa, Black, Jacobs, and
  Malik]{kanazawa2018end}
Angjoo Kanazawa, Michael~J Black, David~W Jacobs, and Jitendra Malik.
\newblock End-to-end recovery of human shape and pose.
\newblock In \emph{Proceedings of the IEEE conference on computer vision and
  pattern recognition}, pages 7122--7131, 2018.

\bibitem[Kingma and Ba(2014)]{kingma2014adam}
Diederik~P Kingma and Jimmy Ba.
\newblock {Adam: A Method for Stochastic Optimization}.
\newblock In \emph{ICLR}, 2014.

\bibitem[Kocabas et~al.(2020)Kocabas, Athanasiou, and Black]{kocabas2020vibe}
Muhammed Kocabas, Nikos Athanasiou, and Michael~J Black.
\newblock {VIBE: Video Inference for Human Body Pose and Shape Estimation}.
\newblock In \emph{CVPR}, pages 5253--5263, 2020.

\bibitem[Kocabas et~al.(2021)Kocabas, Huang, Hilliges, and
  Black]{kocabas2021pare}
Muhammed Kocabas, Chun-Hao~P Huang, Otmar Hilliges, and Michael~J Black.
\newblock Pare: Part attention regressor for 3d human body estimation.
\newblock In \emph{Proceedings of the IEEE/CVF International Conference on
  Computer Vision}, pages 11127--11137, 2021.

\bibitem[Lin et~al.(2021)Lin, Wang, and Liu]{lin2021end}
Kevin Lin, Lijuan Wang, and Zicheng Liu.
\newblock End-to-end human pose and mesh reconstruction with transformers.
\newblock In \emph{CVPR}, pages 1954--1963, 2021.

\bibitem[Liu et~al.(2021)Liu, Chen, Feng, Wu, Ji, Yang, and Wang]{liu2021deep}
Zhenguang Liu, Haoming Chen, Runyang Feng, Shuang Wu, Shouling Ji, Bailin Yang,
  and Xun Wang.
\newblock Deep dual consecutive network for human pose estimation.
\newblock In \emph{Proceedings of the IEEE/CVF Conference on Computer Vision
  and Pattern Recognition}, pages 525--534, 2021.

\bibitem[Loper et~al.(2015)Loper, Mahmood, Romero, Pons-Moll, and
  Black]{loper2015smpl}
Matthew Loper, Naureen Mahmood, Javier Romero, Gerard Pons-Moll, and Michael~J
  Black.
\newblock Smpl: A skinned multi-person linear model.
\newblock \emph{ACM transactions on graphics (TOG)}, 34\penalty0 (6):\penalty0
  1--16, 2015.

\bibitem[Martinez et~al.(2017)Martinez, Hossain, Romero, and
  Little]{martinez2017simple}
Julieta Martinez, Rayat Hossain, Javier Romero, and James~J Little.
\newblock A simple yet effective baseline for 3d human pose estimation.
\newblock In \emph{ICCV}, pages 2640--2649, 2017.

\bibitem[Mayer et~al.(2016)Mayer, Ilg, Hausser, Fischer, Cremers, Dosovitskiy,
  and Brox]{mayer2016large}
Nikolaus Mayer, Eddy Ilg, Philip Hausser, Philipp Fischer, Daniel Cremers,
  Alexey Dosovitskiy, and Thomas Brox.
\newblock A large dataset to train convolutional networks for disparity,
  optical flow, and scene flow estimation.
\newblock In \emph{Proceedings of the IEEE conference on computer vision and
  pattern recognition}, pages 4040--4048, 2016.

\bibitem[Moon and Lee(2020)]{moon2020i2l}
Gyeongsik Moon and Kyoung~Mu Lee.
\newblock I2l-meshnet: Image-to-lixel prediction network for accurate 3d human
  pose and mesh estimation from a single rgb image.
\newblock In \emph{European Conference on Computer Vision}, pages 752--768.
  Springer, 2020.

\bibitem[Moreno-Noguer(2017)]{moreno20173d}
Francesc Moreno-Noguer.
\newblock 3d human pose estimation from a single image via distance matrix
  regression.
\newblock In \emph{Proceedings of the IEEE Conference on Computer Vision and
  Pattern Recognition}, pages 2823--2832, 2017.

\bibitem[Osman et~al.(2020)Osman, Bolkart, and Black]{STAR:2020}
Ahmed A~A Osman, Timo Bolkart, and Michael~J. Black.
\newblock {STAR}: A sparse trained articulated human body regressor.
\newblock In \emph{European Conference on Computer Vision (ECCV)}, pages
  598--613, 2020.
\newblock URL \url{https://star.is.tue.mpg.de}.

\bibitem[Pavlakos et~al.(2017)Pavlakos, Zhou, Derpanis, and
  Daniilidis]{pavlakos2017coarse}
Georgios Pavlakos, Xiaowei Zhou, Konstantinos~G Derpanis, and Kostas
  Daniilidis.
\newblock Coarse-to-fine volumetric prediction for single-image 3d human pose.
\newblock In \emph{Proceedings of the IEEE conference on computer vision and
  pattern recognition}, pages 7025--7034, 2017.

\bibitem[Pavlakos et~al.(2019)Pavlakos, Choutas, Ghorbani, Bolkart, Osman,
  Tzionas, and Black]{SMPL-X:2019}
Georgios Pavlakos, Vasileios Choutas, Nima Ghorbani, Timo Bolkart, Ahmed A.~A.
  Osman, Dimitrios Tzionas, and Michael~J. Black.
\newblock Expressive body capture: {3D} hands, face, and body from a single
  image.
\newblock In \emph{Proceedings IEEE Conf. on Computer Vision and Pattern
  Recognition (CVPR)}, pages 10975--10985, 2019.

\bibitem[Pfister et~al.(2015)Pfister, Charles, and
  Zisserman]{pfister2015flowing}
Tomas Pfister, James Charles, and Andrew Zisserman.
\newblock Flowing convnets for human pose estimation in videos.
\newblock In \emph{Proceedings of the IEEE international conference on computer
  vision}, pages 1913--1921, 2015.

\bibitem[Ranjan and Black(2017)]{ranjan2017optical}
Anurag Ranjan and Michael~J Black.
\newblock Optical flow estimation using a spatial pyramid network.
\newblock In \emph{Proceedings of the IEEE conference on computer vision and
  pattern recognition}, pages 4161--4170, 2017.

\bibitem[Ranjan et~al.(2018)Ranjan, Romero, and Black]{ranjan2018learning}
Anurag Ranjan, Javier Romero, and Michael~J Black.
\newblock Learning human optical flow.
\newblock \emph{arXiv preprint arXiv:1806.05666}, 2018.

\bibitem[Ranjan et~al.(2020)Ranjan, Hoffmann, Tzionas, Tang, Romero, and
  Black]{ranjan2020learning}
Anurag Ranjan, David~T Hoffmann, Dimitrios Tzionas, Siyu Tang, Javier Romero,
  and Michael~J Black.
\newblock Learning multi-human optical flow.
\newblock \emph{International Journal of Computer Vision}, 128\penalty0
  (4):\penalty0 873--890, 2020.

\bibitem[Stenum et~al.(2021)Stenum, Cherry-Allen, Pyles, Reetzke, Vignos, and
  Roemmich]{stenum2021applications}
Jan Stenum, Kendra~M Cherry-Allen, Connor~O Pyles, Rachel~D Reetzke, Michael~F
  Vignos, and Ryan~T Roemmich.
\newblock Applications of pose estimation in human health and performance
  across the lifespan.
\newblock \emph{Sensors}, 21\penalty0 (21):\penalty0 7315, 2021.

\bibitem[Sun et~al.(2018)Sun, Yang, Liu, and Kautz]{Sun2018PWC-Net}
Deqing Sun, Xiaodong Yang, Ming-Yu Liu, and Jan Kautz.
\newblock {PWC-Net}: {CNNs} for optical flow using pyramid, warping, and cost
  volume.
\newblock In \emph{CVPR}, 2018.

\bibitem[Teed and Deng(2020)]{teed2020raft}
Zachary Teed and Jia Deng.
\newblock Raft: Recurrent all-pairs field transforms for optical flow.
\newblock In \emph{European conference on computer vision}, pages 402--419.
  Springer, 2020.

\bibitem[Tran et~al.(2016)Tran, Bourdev, Fergus, Torresani, and
  Paluri]{tran2016deep}
Du~Tran, Lubomir Bourdev, Rob Fergus, Lorenzo Torresani, and Manohar Paluri.
\newblock Deep end2end voxel2voxel prediction.
\newblock In \emph{Proceedings of the IEEE conference on computer vision and
  pattern recognition workshops}, pages 17--24, 2016.

\bibitem[Truong et~al.(2020)Truong, Danelljan, and Timofte]{GLUNet_Truong_2020}
Prune Truong, Martin Danelljan, and Radu Timofte.
\newblock {GLU-Net}: Global-local universal network for dense flow and
  correspondences.
\newblock In \emph{CVPR}, 2020.

\bibitem[Ullah et~al.(2018)Ullah, Muhammad, Del~Ser, Baik, and
  de~Albuquerque]{ullah2018activity}
Amin Ullah, Khan Muhammad, Javier Del~Ser, Sung~Wook Baik, and Victor Hugo~C
  de~Albuquerque.
\newblock Activity recognition using temporal optical flow convolutional
  features and multilayer lstm.
\newblock \emph{IEEE Transactions on Industrial Electronics}, 66\penalty0
  (12):\penalty0 9692--9702, 2018.

\bibitem[Ulyanov et~al.(2018)Ulyanov, Vedaldi, and Lempitsky]{ulyanov2018deep}
Dmitry Ulyanov, Andrea Vedaldi, and Victor Lempitsky.
\newblock Deep image prior.
\newblock In \emph{Proceedings of the IEEE conference on computer vision and
  pattern recognition}, pages 9446--9454, 2018.

\bibitem[Vaswani et~al.(2017)Vaswani, Shazeer, Parmar, Uszkoreit, Jones, Gomez,
  Kaiser, and Polosukhin]{Vaswani17}
Ashish Vaswani, Noam Shazeer, Niki Parmar, Jakob Uszkoreit, Llion Jones,
  Aidan~N Gomez, \L~ukasz Kaiser, and Illia Polosukhin.
\newblock {Attention is All you Need}.
\newblock In \emph{NIPS}, 2017.

\bibitem[von Marcard et~al.(2018)von Marcard, Henschel, Black, Rosenhahn, and
  Pons-Moll]{von2018recovering}
Timo von Marcard, Roberto Henschel, Michael~J Black, Bodo Rosenhahn, and Gerard
  Pons-Moll.
\newblock Recovering accurate 3d human pose in the wild using imus and a moving
  camera.
\newblock In \emph{Proceedings of the European Conference on Computer Vision
  (ECCV)}, pages 601--617, 2018.

\bibitem[Yu et~al.(2021)Yu, Ni, Xu, Wang, Zhao, and Zhang]{yu2021towards}
Zhenbo Yu, Bingbing Ni, Jingwei Xu, Junjie Wang, Chenglong Zhao, and Wenjun
  Zhang.
\newblock Towards alleviating the modeling ambiguity of unsupervised monocular
  3d human pose estimation.
\newblock In \emph{Proceedings of the IEEE/CVF International Conference on
  Computer Vision}, pages 8651--8660, 2021.

\bibitem[Zhao et~al.(2017)Zhao, Wang, and Martinez]{zhao2017simple}
Ruiqi Zhao, Yan Wang, and Aleix~M Martinez.
\newblock A simple, fast and highly-accurate algorithm to recover 3d shape from
  2d landmarks on a single image.
\newblock \emph{IEEE transactions on pattern analysis and machine
  intelligence}, 40\penalty0 (12):\penalty0 3059--3066, 2017.

\bibitem[Zou et~al.(2021)Zou, Guo, Zuo, Wang, Wang, Hu, Chen, Gong, and
  Cheng]{zou2021eventhpe}
Shihao Zou, Chuan Guo, Xinxin Zuo, Sen Wang, Pengyu Wang, Xiaoqin Hu, Shoushun
  Chen, Minglun Gong, and Li~Cheng.
\newblock Eventhpe: Event-based 3d human pose and shape estimation.
\newblock In \emph{Proceedings of the IEEE/CVF International Conference on
  Computer Vision}, pages 10996--11005, 2021.

\end{thebibliography}
\clearpage
\appendix

\setcounter{page}{1}

{
\centering
\Large
\textbf{Bootstrapping Human Optical Flow and Pose} \\
\vspace{0.5em}Supplementary Material \\
\vspace{1.0em}
}

\appendix

We provide additional results excluded from the main paper due to spatial constraints.

\section{Ablation study}
\label{sec:ablation}

To motivate the design choices of our method we conduct an ablation study.

\paragraph{Loss terms in pose optimization.}
We first examine how the pose estimation performance of our method varies as we add new loss terms in \Eq{2}.
We report these results in \Table{ablation}, using the Human3.6M dataset.
As shown, each loss term contributes to enhanced MPJPE, demonstrating that all loss components are important.
We emphasize once more here that, while there are four loss terms in total, we use the \emph{same} hyperparameter setting for all our experiments.

\begin{table}[h!]
\begin{center}
    \setlength{\tabcolsep}{6pt}
    \begin{tabular}{@{}lc@{}} 
     \toprule
     Method & MPJPE $\downarrow$ \\
      \midrule
     Initial pose estimates (METRO)  & 54.07\\
     $\loss{3D}$ & 54.07\\
     $\loss{3D}$ $+$ $\loss{2D}$ & 53.93\\
     $\loss{3D}$ $+$ $\loss{2D}$ $+$ $\loss{temp}$ (without bone consistency) & 53.45\\
     $\loss{3D}$ $+$ $\loss{2D}$ $+$ $\loss{temp}$  & 53.29\\
     $\loss{3D}$ $+$ $\loss{2D}$ $+$ $\loss{temp}$ $+$ $\loss{opt}$  & {\bf 53.15}\\
     \bottomrule
    \end{tabular}
\end{center}
\caption{{\bf Ablation study on pose-related loss terms -- }
Ablation study on the Human3.6M dataset showing the effects of adding different loss terms to our pose refinement pipeline. 
All loss terms contribute to the enhancement of human pose accuracy.
}%
\label{tab:ablation}
\end{table}

\begin{figure}[h!]
\centering
\subfigure[MPJPE vs \#cycles]{\includegraphics[width=0.24\linewidth, trim= 0 0 40 35, clip]{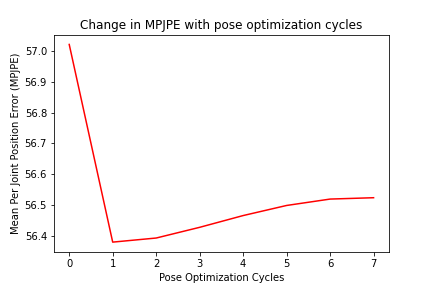}}
\subfigure[EPE vs \#cycles]{\includegraphics[width=0.24\linewidth, trim= 0 0 40 35, clip]{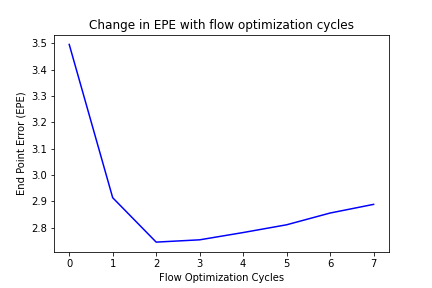}}
\subfigure[MPJPE vs \#cycles /w GT]{\includegraphics[width=0.24\linewidth, trim= 0 0 40 35, clip]{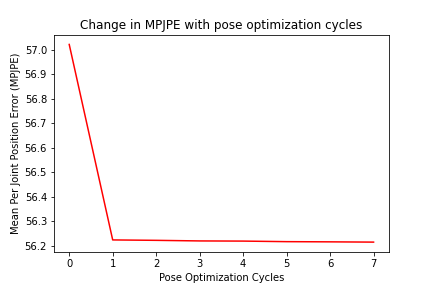}}
\subfigure[EPE vs \#cycles /w GT]{\includegraphics[width=0.24\linewidth, trim= 0 0 40 35, clip]{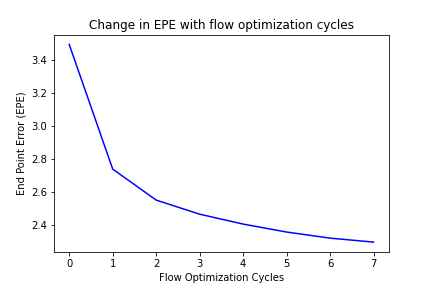}}

\caption{
{\bf Ablation study on the number of cycles -- }
MPJPE and EPE change with respect to the number of optimization cycles, on a video sequence of the Human3.6M dataset. 
(a--b) when the optimization is purely based on our method using estimated pose and flow, and (c--d) when we replace the optimization cycle with ground-truth measurements rather than estimated pose and flow.
For the case when the estimated pose and flow are used, the best pose is already achieved at the first optimization cycle, whereas the flow at the second.
The results then deteriorate, showing `drifting'.
When using the ground truth, this does not happen, further suggesting drifting.
Nonetheless, optimized results with our method improve over initial estimates even with drifting.
While these measurements are for a single video, the same trend can be observed in general.}

\label{fig:ablation_cycles}
\end{figure}

\paragraph{Number of optimization cycles}
With a representative video from the Human3.6M dataset, we report how pose estimation accuracy and human optical flow accuracy change as we perform more optimization cycles.
As shown in \Figure{ablation_cycles}~(a--b), the best pose is achieved after the first pose optimization cycle, whereas the best flow is achieved after the second.
This demonstrates that a single pose optimization cycle is enough to take optical flow into account when estimating the poses, which can then correct optical flow with the enhanced poses.
The increase in error afterward indicates that there is a potential drift after more optimization cycles.
For example, when too many optimization cycles are performed, the RAFT network can overfit to the rough flow estimates shown in \Figure{flow_sketch}~(c), which \emph{will} contain errors, leading to degradation.

As shown in 
\Figure{ablation_cycles}~(c--d), where these estimation errors are not present, this drifting does not happen.
We note that this is a limitation of \emph{any} pipeline that self-boot-straps and is not unique to our method.
In addition, even when degradation happens, the performance still is much better than the initial MPJPE and EPE values.
Based on these results, we perform one pose optimization cycle and two flow cycles for all our experiments.

\end{document}